\documentclass{article}

\usepackage{amsmath}
\usepackage{microtype}
\usepackage{graphicx}
\usepackage{subfigure}
\usepackage{float}
\usepackage{booktabs} % for professional tables
\usepackage{etoolbox}
\usepackage{enumitem}
\usepackage[accepted]{icml2021}
\usepackage{hyperref}

\icmltitlerunning{Geometrical Homogeneous Clustering}

\begin{document}

\twocolumn[
\icmltitle{Geometrical Homogeneous Clustering for Image Data Reduction}

% It is OKAY to include author information, even for blind
% submissions: the style file will automatically remove it for you
% unless you've provided the [accepted] option to the icml2021
% package.

% List of affiliations: The first argument should be a (short)
% identifier you will use later to specify author affiliations
% Academic affiliations should list Department, University, City, Region, Country
% Industry affiliations should list Company, City, Region, Country

% You can specify symbols, otherwise they are numbered in order.
% Ideally, you should not use this facility. Affiliations will be numbered
% in order of appearance and this is the preferred way.
\icmlsetsymbol{equal}{*}

\begin{icmlauthorlist}
\icmlauthor{Shril Mody}{equal,inst}
\icmlauthor{Janvi Thakkar}{equal,inst}
\icmlauthor{Devvrat Joshi}{equal,inst}
\icmlauthor{Siddharth Soni}{equal,inst}
\icmlauthor{Rohan Patil}{inst}
\icmlauthor{Nipun Batra}{inst}

\end{icmlauthorlist}

\icmlaffiliation{inst}{IIT Gandhinagar}

\icmlcorrespondingauthor{Shril Mody}{paresh.mody@iitgn.ac.in}
% \icmlcorrespondingauthor{Janvi}{janvi.thakkar@iitgn.ac.in}
% \icmlcorrespondingauthor{Devvrat}{devvrat.joshi@iitgn.ac.in}
\icmlcorrespondingauthor{Siddharth Soni}{siddharth.soni@iitgn.ac.in}
% \icmlcorrespondingauthor{Nipun}{nipun.batra@iitgn.ac.in}
% \icmlcorrespondingauthor{Rohan}{rohan.patil@iitgn.ac.in}
% You may provide any keywords that you
% find helpful for describing your paper; these are used to populate
% the "keywords" metadata in the PDF but will not be shown in the document
\icmlkeywords{Machine Learning, ICML}

\vskip 0.3in
]

% this must go after the closing bracket ] following \twocolumn[ ...

% This command actually creates the footnote in the first column
% listing the affiliations and the copyright notice.
% The command takes one argument, which is text to display at the start of the footnote.
% The \icmlEqualContribution command is standard text for equal contribution.
% Remove it (just {}) if you do not need this facility.

%\printAffiliationsAndNotice{}  % leave blank if no need to mention equal contribution
\printAffiliationsAndNotice{\icmlEqualContribution} 

% otherwise use the standard text.

\begin{abstract}
In this paper, we present novel variations of an earlier approach called homogeneous clustering algorithm for reducing dataset size. The intuition behind the approaches proposed in this paper is to partition the dataset into homogeneous clusters and select some images which contribute significantly to the accuracy. Selected images are the proper subset of the training data and thus are human-readable. We propose four variations upon the baseline algorithm-RHC. The intuition behind the first approach, RHCKON, is that the boundary points contribute significantly towards the representation of clusters. It involves selecting $k$ farthest and one nearest neighbour of the centroid of the clusters. In the following two approaches (KONCW and CWKC), we introduce the concept of cluster weights. They are based on the fact that larger clusters contribute more than smaller sized clusters. The final variation is GHCIDR which selects points based on the geometrical aspect of data distribution.
We performed the experiments on two deep learning models- Fully Connected Networks (FCN) and VGG1. We experimented with the four variants on three datasets- MNIST, CIFAR10, and Fashion-MNIST. We found that GHCIDR gave the best accuracy of 99.35\%, 81.10\%, and 91.66\% and a training data reduction of 87.27\%, 32.34\%, and 76.80\% on MNIST, CIFAR10, and Fashion-MNIST respectively.
\end{abstract}
% mention that various approaches for dataset reduction have been approached in the past, we pick up one such approach which is shown to be promising RHC. Explain in 1-2 line RHC, next para limitations of RHC 2-3 lines, next para tell how we improve the performance of RHC
\section{Introduction}

Various approaches for dataset reduction have been proposed in the past. One such popular technique is called homogeneous clustering (RHC)~\cite{ougiaroglou2012efficient}. RHC clusters the given dataset by considering the labels of each data point. The dataset is partitioned into homogeneous clusters; i.e. all the data points belonging to that cluster will have the same label. The final reduced dataset is constructed by taking the centroids of these homogeneous clusters.

The RHC algorithm only includes the centroid of the homogeneous clusters in the reduced set. It disregards the importance of other images which forms the cluster boundary. Boundary points can potentially play a significant role in distinguishing the labels of neighbouring clusters. The varying size of the clusters is also not taken into account by RHC. Larger clusters contain more images, and so their contribution should be more. Our work attempts to further improve the idea of RHC by taking into account the geometric features involved in generating clusters. We focused on reducing the dataset such that it stays human readable. Human-readable data can be used to interpret the results in case of unexpected output easily.

We propose four variations of our approach. All the algorithms differ in the selection of data points after the creation of homogeneous clusters. The first approach, RHCKON focuses on picking the data points greedily near the centroid of the clusters ($k$-farthest and 1-nearest neighbours). The intuition behind RHCKON is that the boundary points are a factor for classification between adjacent clusters~\cite{olvera2010new}. The second variation, KONCW gives weights to each cluster according to their size and selects images based on these weights. Large-sized clusters could contribute more towards the accuracy as they contain more images, so we can include them by giving weights. The next variation, CWKC greedily chooses the images which are evenly spread across the cluster boundary. Here, the number of images chosen from the cluster depends on their weight. The final algorithm, GHCIDR combines all the previous variations by dividing the cluster into annular regions and selecting images from each region. We can get a notion of the entire cluster by selecting images equally spaced from the centroid rather than selecting only the farthest and nearest images from a cluster. 

All these four variants were tested on three image datasets: MNIST (60,000 training images)~\cite{deng2012mnist}, CIFAR10 (50,000 training images)~\cite{krizhevsky2009learning} and Fashion-MNIST (60,000 training images)~\cite{xiao2017fashion}. We tested our reduced datasets on two deep learning models, namely, Fully Connected Networks (FCN~\cite{long2015fully}) and VGG1~\cite{wang2015places205}. We found that all four variants performed better than the baseline RHC algorithm on both models. GHCIDR on VGG1 performs the best among the four and gives approximately the same accuracy as that of the model trained on full dataset.    +

Our code  \footnote{Here is the link to the code \url{https://github.com/SoniSiddharth/GHCIDR}.} is fully reproducible.

\section{Background (Reduction through Homogeneous Clustering)}
A variety of sampling-based algorithms were proposed to reduce the size of the annotated datasets. In this section, we describe one of the previously proposed approaches called ``Reduction through Homogeneous Clustering (RHC)''. RHC focused on constructing homogeneous clusters. Homogeneous clusters contain data points belonging to the same class. The algorithm maintains a set of clusters. Initially, this set consists of the proper training dataset as a single cluster. On each iteration, we pop out a cluster from this set. Now, there are two possible conditions for the clusters: homogeneous \& non-homogeneous. If the cluster is homogeneous, it gets added to the Condensed Set. Condensed Set is the final output of the RHC algorithm. If it is non-homogeneous, then we recursively cluster the points using $K$-means~\cite{chen2009k} algorithm until either all clusters become homogeneous or there is no cluster left. At each step, we initialize $K$-means by the centroids of each class belonging to the cluster. The dataset is reduced by selecting centroids of individual homogeneous clusters. RHC was tested on various annotated datasets ~\cite{alcala2011keel}, namely, Letter recognition (LR), Pen-Digits (PD) and Landsat Satellite (LS). It gave better accuracy with lesser time complexity than classical ML algorithms such as K-NN and  Prototype Selection by Clustering (PSC) ~\cite{olvera2010new}.

\begin{figure}[h]
    \centering
    % \subfloat
    {\includegraphics[width=3.5cm]{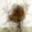}}%
    \qquad
    % \subfloat
    {\includegraphics[width=3.5cm]{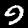}}%
    \caption{Non Human readable images of RHC (left: CIFAR10 (class: bird), right: MNIST (class: 9))}%
    \label{fig:nonhuman}%
    % \vspace{-2}
\end{figure}

The original RHC work primarily focused on reducing the annotated dataset. One of the shortcomings of the RHC algorithm was that it selected centroids of the images that could be non-human readable as shown in Figure \ref{fig:nonhuman}. We modified the RHC algorithm to produce human-readable samples giving high accuracy.

\section{Proposed Approach}
% The RHC algorithm took only the centroid of the images from each of the clusters, and they were also non-human readable. All the clusters are homogeneous. We can improve the RHC algorithm by selecting more images from each cluster to make the final reduced dataset. The main problem that arises now is to select which images from the clusters so that they contribute the most to the final accuracy. 
The RHC algorithm chooses the centroid of each homogeneous cluster. This aggregation of images is not human readable. RHC also ignores the size of clusters and selects a single image (centroid), making the contribution of small and large clusters to be equal in the reduced dataset. These drawbacks are caused by RHC's selection method of images from homogeneous clusters. We propose four different variants which overcome these limitations.

% Our aim is to efficiently select images from clusters in order to maximize the accuracy while minimizing the dataset size required. (IMPORTANT)

%\subsection{Approach 0: Random Selection}

%We can ignore the RHC algorithm and directly select a random $\%$ of images and train them on some standard model. 
%The algorithm is uncertain in terms of accuracy, although being exact in reduction rate. The final model trained using random data could be biased towards certain classes, leading to a defective model.

\subsection{Approach 1: RHC + $K$-Farthest + OneNearest (RHCKON)}

The RHC algorithm does not select the boundary images from each of the homogeneous clusters, which can potentially be the distinguishing factor for classification. 
Boundary points of a cluster 
Most of the boundary points have properties similar to more than one cluster, so classifying them correctly can have a considerable impact on test accuracy.

\begin{figure}[h]
  \includegraphics[width=\linewidth , height=2in]{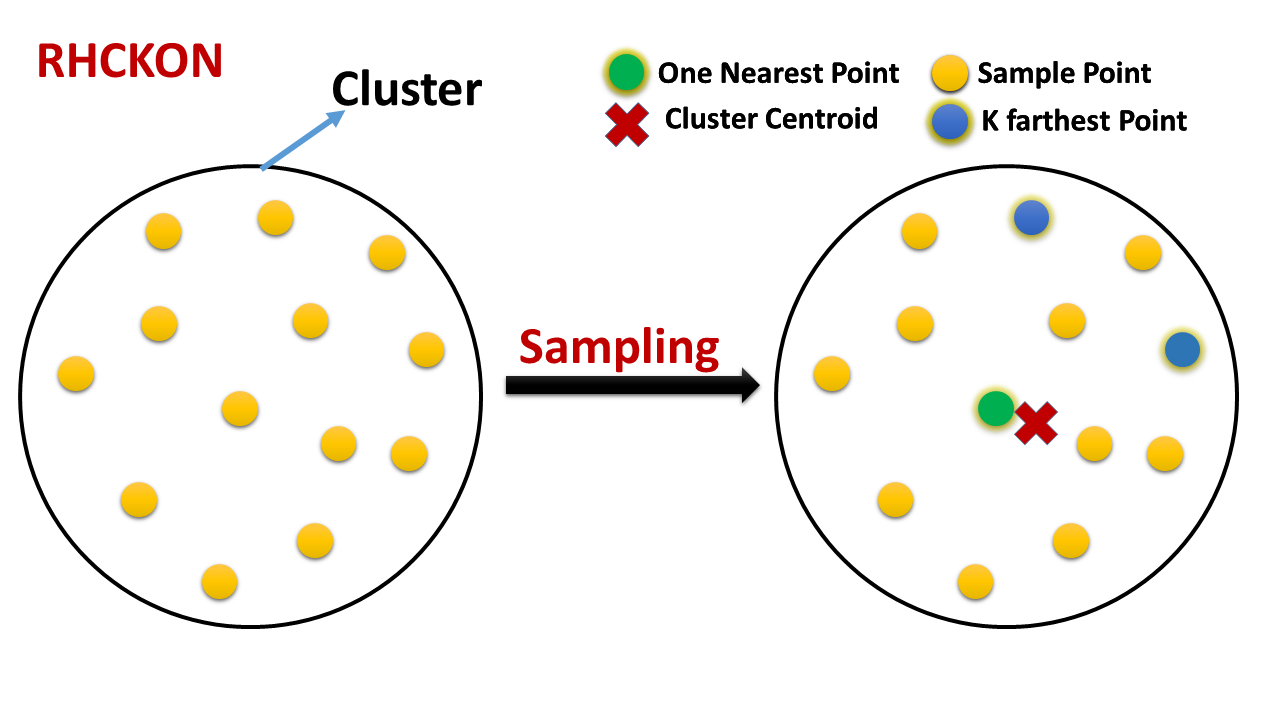}
  \caption{Approach 1: RHCKON. This image shows one of the many homogeneous clusters. We sample the point closest to the centroid and K farthest points from the centroid. Choosing boundary points may provide diverse images to achieve better generalization.}
%   \Description{RHCKON algorithm}
  \label{fig:rhckon.PNG}
\end{figure}

\textbf{Algorithm:} In RHCKON (Figure ~\ref{fig:rhckon.PNG}), we choose one nearest point and $K$ farthest points from the centroid of each cluster. Here, these $K$ points represent the boundary points for each cluster. 

In RHCKON, we envision the following three main drawbacks.
\begin{enumerate}[noitemsep,topsep=0pt]
    \item It may be prone to outliers (more boundary points in the reduced set).
    \item Reduction is less in comparison to RHC because of more images selected from each cluster.
    \item RHCKON selects the same number of points from each cluster. Therefore, it could lead to more fraction of images from smaller sized clusters in the reduced set.
    % \item Reduction is less because we will sample more points, unlike RHC where we were picking the only centroid now we are picking some more points from each cluster.
    % %\item Reduction rate is not in our control and changes with the value of K.
\end{enumerate}

\begin{figure}[htp]
  \includegraphics[width=\linewidth]{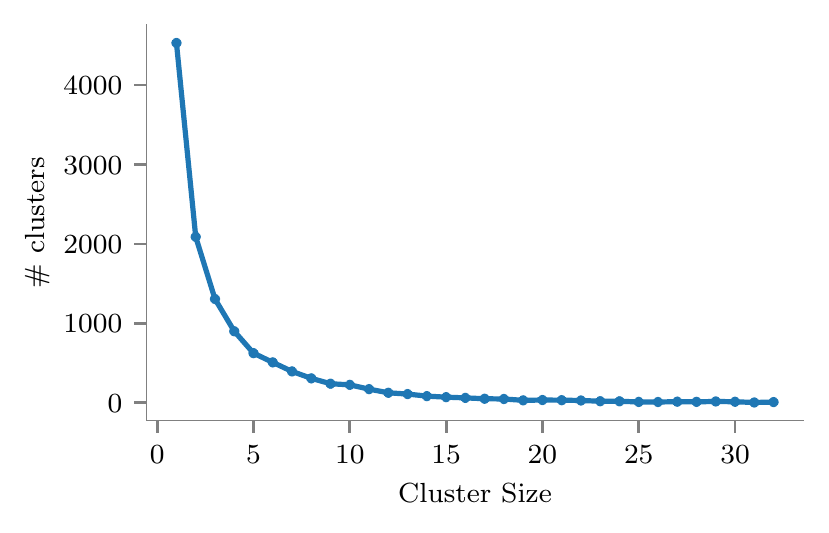}
  \caption{No. of Clusters vs. Cluster Size. Clusters with small size are large in number, and their count decreases exponentially with their size. 10000 smaller clusters contain 15000 images whereas 2400 larger clusters contain the remaining 35000 images.}
%   \Description{Cluster Size}
  \label{fig:cluster.PNG}
%   \vspace{-2}
\end{figure}

To illustrate the point 3 above, we plot the size of clusters versus cluster count on CIFAR10 dataset in Figure \ref{fig:cluster.PNG}  
%As the cluster size increases, their number decreases almost exponentially.
The clusters of smaller size are significantly large in number, whereas clusters of larger size are small in number. Out of 12,400 clusters, nearly 10,000 clusters lie in a size range between 1 and 5. However, the significant chunk of data points lies in the large-sized clusters. Thus, by selecting one nearest and $K$ farthest samples, we are giving the same weightage to all the clusters irrespective of their size.

%figure 3: vs their count Make this plot nicely, using the article shown from my website, please follow this article. What is the main takeaway from the figure THere are a large clusters with smaller size, how does it impact, one or two lines should be enough

\subsubsection{\textbf{Approach 2 - RHCKON+Cluster Weightage (KONCW)}}

\begin{figure}[]
  \includegraphics[width=\linewidth , height=2in]{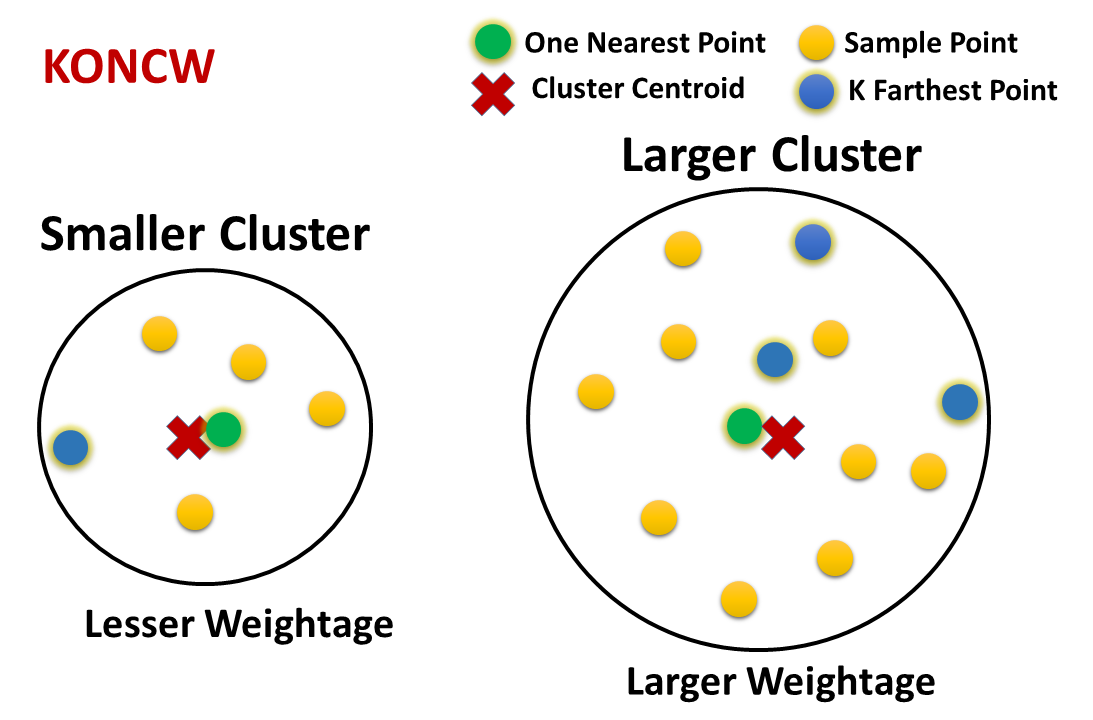}
  \caption{Approach 2: KONCW. This image shows two of the many homogeneous clusters. One of the clusters has fewer images as compared to the other. The number of farthest points selected from the larger cluster is more as it has higher weightage. The intuition is to give importance to a cluster proportional to its size.}
%   \Description{koncw}
  \label{fig:koncw.PNG}
%   \vspace{-2}
\end{figure}

In our second approach (shown in Figure \ref{fig:koncw.PNG}) instead of selecting an equal number of images via RHCKON, we can give a weightage (fraction of selected images) to each cluster according to their size. Thus, the largest cluster will have the highest weightage and vice versa.

\textbf{Algorithm:} We need to select at least one image from each cluster to have a representation of these isolated clusters in our reduced dataset. 
\begin{itemize}[noitemsep,topsep=0pt]
    \item Select a fraction of reduction $(\alpha)$, i.e. select from each cluster a fraction of points $(1-\alpha)$.
    \item $\max((1-\alpha)\times(\textrm{size of cluster}), 1)$ is the number of images selected from each cluster.
    \item Let the number of the images from the $i^{th}$ cluster be $X_i$.
    \item Select one nearest image and, $X_i - 1$ farthest images from the the $i^{th}$ cluster.
\end{itemize}
% We select a fraction of reduction $(\alpha)$, i.e. select from each cluster a fraction of points $(1-\alpha)$. Now, $max((1-\alpha)\times(size of cluster), 1)$ is the number of images selected from each cluster. We need to select atleast one image from each cluster. Let the number of the images from $cluster_i$ be $X_i$. Select one nearest image and, $X_i - 1$ farthest images from the $cluster_i$.
% \begin{algorithm}
% \caption{KONCW Algorithm}\label{koncw_algo}
% \begin{algorithmic}[1]
% \Procedure{KONCW}{}
% \State Select a fraction of reduction $(\alpha)$
% \State From each cluster, select $(1-\alpha)$ fraction of points
% \State $\#images =  max((1-\alpha)\times(size of cluster), 1)$

% \State Let images in $cluster_i$  = $X_i$
% \State Select one nearest image and, $X_i - 1$ farthest images from the $cluster_i$.
% \EndProcedure
% \end{algorithmic}
% \end{algorithm}

However, we envision the following limitations.

\begin{enumerate}[noitemsep,,topsep=0pt]
    \item Larger clusters have higher weightage in the reduced set. Due to this, the fraction of boundary points selected from these clusters increases with the size of clusters which increases the probability of outliers in our reduced dataset.
    %\item We are taking only the farthest images from centroid for all clusters. Larger clusters have higher weightage and will have more images from the farthest set. There is a reasonable probability that the farthest point might be an \textbf{outlier}, and selecting many farthest points makes data more prone to outliers.
    \item Assume an N-dimensional ball representing the cluster of data points. We are choosing the farthest points irrespective of the direction from the centroid of the ball/cluster. Thus, there is a considerable probability that the selected farthest points might be in the \textbf{same direction} from the centroid of the cluster.
\end{enumerate}
\subsubsection{\textbf{Approach 3 - KONCW+K-Centre (CWKC)}}

\begin{figure}[h]
  \includegraphics[width=\linewidth , height=2in]{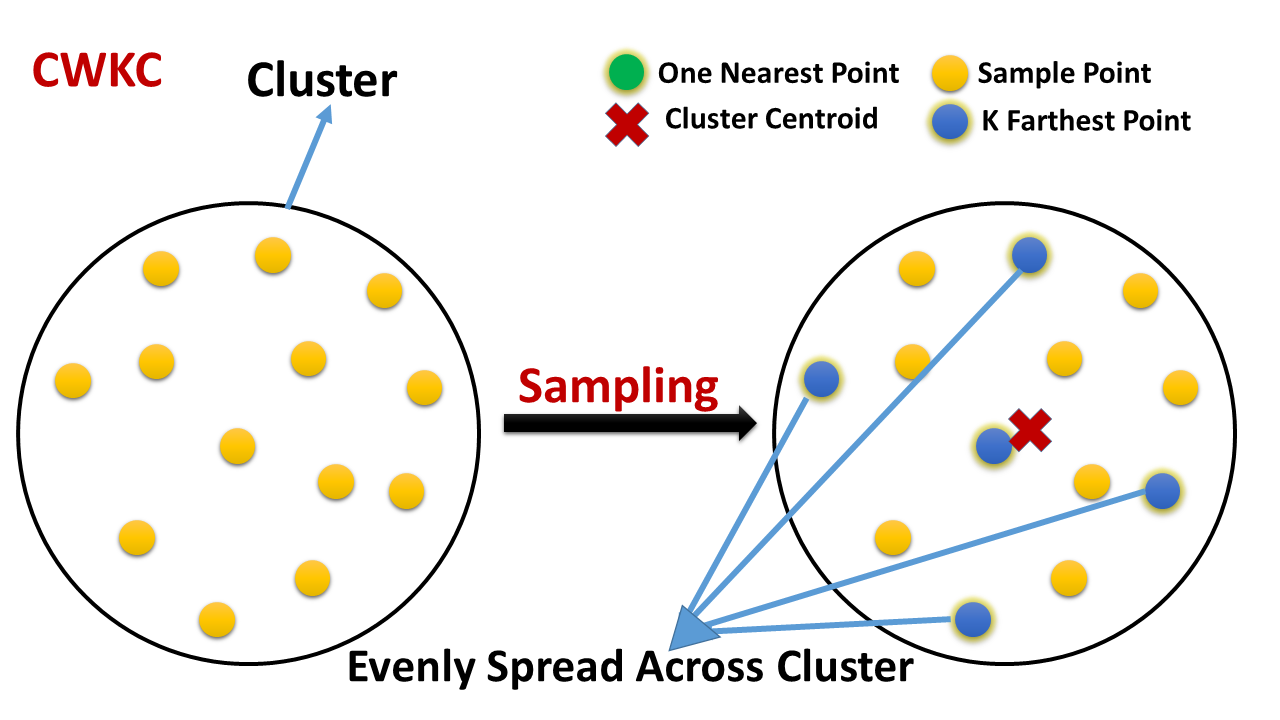}
  \caption{{Approach 3: CWKC. This image shows one of the homogeneous cluster. First, we add the datapoint closest to the centroid. Now, select points which are evenly spread across the cluster boundary. The intuition is to uniformly cover the cluster boundary.}}
%   \Description{cwkc}
  \label{fig:cwkc.PNG}
\end{figure}

We need to select points in different directions from the cluster's centroid. Figure \ref{fig:cwkc.PNG} illusrtaes the CWKC approach.

\textbf{Algorithm:} 
\begin{itemize}[noitemsep,,topsep=0pt]
    \item Initialize the set of centres with one nearest and one farthest point.
    \item Choose the next point, which is farthest from all the points in the set of centres.
    \item It is equivalent to finding a point in the cluster whose minimum distance from all the points in the set is maximum.
    \item Continue extending the set until its size is less than the allowed points from the cluster (weightage).
    \item Apply the same algorithm on all the clusters, and we get our condensed dataset.
\end{itemize}
% Initialize the set of centres with one nearest and one farthest point, then choose the next point, which is farthest from all the points in the set of centres. It is equivalent to finding a point in the cluster whose minimum distance from all the points in the set is maximum. Continue extending the set until its size is less than the allowed points from the cluster (weightage). Apply the same algorithm on all the clusters, and we get our condensed dataset.

%The accuracy of CWKC improved a little but not much in comparison to KONCW. \textbf{Fig. \ref{fig:cwkc.PNG}}.

%The reason for outliers experimentally holds for this method. Even after choosing different directions, we are not getting improved results from that choosing the same directions.

%% plots 

\subsubsection{\textbf{Approach 4 - Geometrical Homogeneous Clustering for Image Data Reduction (GHCIDR)}}
\begin{figure}[h]
  \includegraphics[width=\linewidth , height=2in]{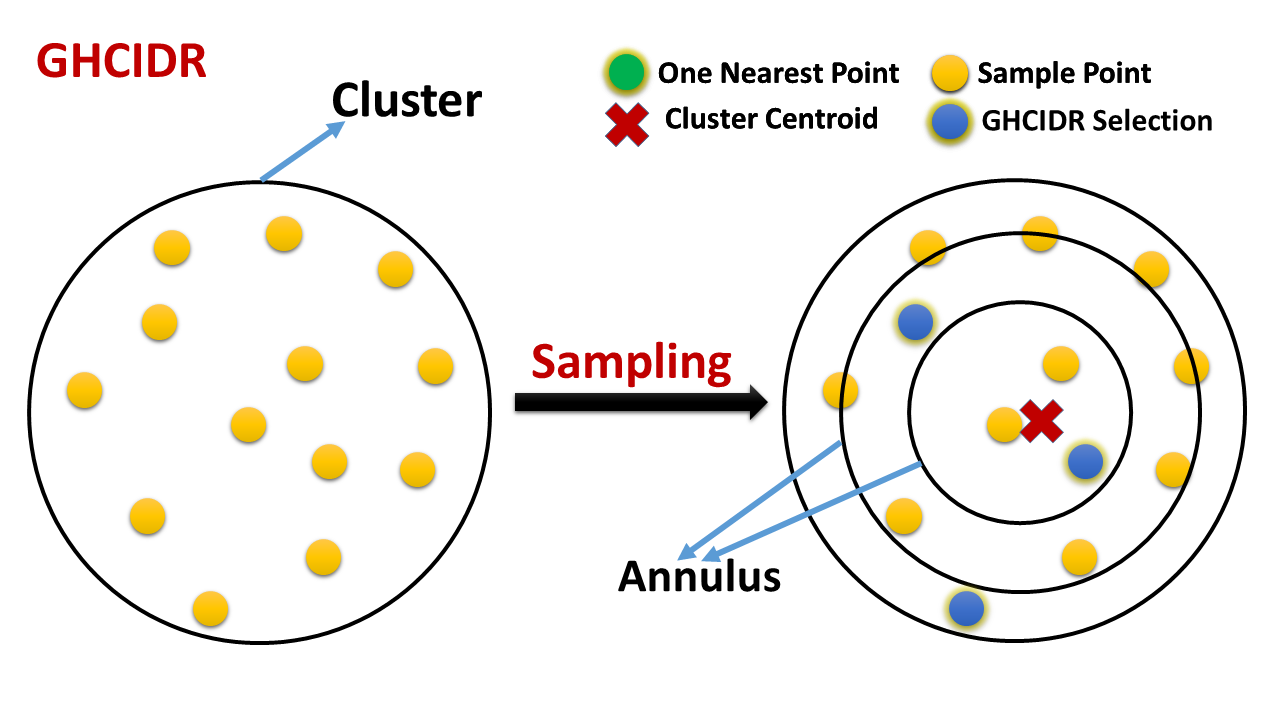}
  \caption{Approach 4: GHCIDR. This is one of the homogeneous clusters. Divide the cluster into different annular regions and select the points nearest to the average distance of the corresponding annulus. The intuition is to select points from the entire points of the cluster.}
%   \Description{Annulus}
  \label{fig:annulus.PNG}
\end{figure}

Selecting the nearest point from the centroid makes a notion of \textbf{centrality} that the given point represents the entire cluster. In our final approach called GHCIDR (shown in Figure \ref{fig:annulus.PNG}) we choose images uniformly from the entire volume of the cluster. This way, we can encompass all the cluster features.

% Taking images from the entire volume of the cluster and still satisfying the centrality is an important aspect in designing the reduction algorithm.
% So, we can take images spread over the entire cluster to encapsulate all the cluster features.

\textbf{Algorithm:} Consider a cluster as a ball of $N$ dimensions.
\begin{itemize}[noitemsep,,topsep=0pt]
    \item Let $maxDist$ be the distance of the farthest point from the centroid.
    \item Divide $maxDist$ into partitions of equal size to get $(maxDist/\beta)$ annuluses of the ball, with the innermost part being a sphere.
    \item $\beta$ depends on the weightage of the cluster and the reduction rate $\alpha$.
    \item For each annulus $i$, let the inner radius and outer radius of the annulus be $R1_i$ and $R2_i$.
    \item Find the point belonging to annulus $i$ and having distance from the centroid to be $(R1_i + R2_i)/2$.
    \item Put that point along with the one nearest point from the central sphere into the condensed set.
\end{itemize}
% \textbf{Algorithm:} Consider a cluster as a ball of N dimensions. Let $maxDist$ be the distance of the farthest point from the centroid. Now, divide maxDist into partitions of equal size to get $(maxDist/\beta)$ annuluses of the ball, with the innermost part being a sphere as shown in \textbf{Figure: \ref{fig:annulus.PNG}}. $\beta$ depends on the weightage of the cluster and the reduction rate. For each annulus $i$, let the inner radius and outer radius of the annulus be $R1_i$ and $R2_i$. Next, find the point which belongs to annulus $i$ and having distance from the centroid to be closest to $(R1_i + R2_i)/2$. Put that point along with the one nearest point from the central sphere into the condensed set.

We set $\beta$ to get exactly one point from each annulus. 
$$\beta = \frac{maxLength}{((1-\alpha)\times(size \,of \,cluster))}$$

% \begin{align*}
%   \beta = \frac{maxLength}{((1-\alpha)\times(size \,of \,cluster))}\\
% \end{align*}
% There might be some annulus $i$ that does not have any data point. To handle this case, select the point from the annulus $i+1$ (outer), which has a distance from the centroid closest to $r2_i$. It is the same as choosing the nearest point to annulus $i$ from annulus $i+1$. Selecting the nearest point to annulus $i$ makes sense because, in the test set, there might be some point that would have fallen in the empty annulus. This ensures that even these data points get correctly classified, and the condensed set does not miss them.
%% plots

\begin{table*}[]
\centering
\begin{tabular}{|l|r|r|r|r|r|r|}
\hline & \multicolumn{6}{c|}{\textbf{FCN}}                                  \\ \hline & \multicolumn{3}{c|}{\textbf{MNIST}}  & \multicolumn{3}{c|}{\textbf{FMNIST}} \\ \hline
Algorithm & \multicolumn{1}{l|}{\begin{tabular}[c]{@{}l@{}}\%Accuracy on \\ reduced data\end{tabular}} & \multicolumn{1}{l|}{\%Reduction} & \multicolumn{1}{l|}{\begin{tabular}[c]{@{}l@{}}\%Accuracy on\\    random data\end{tabular}} & \multicolumn{1}{l|}{\begin{tabular}[c]{@{}l@{}}\%Accuracy on \\    reduced data\end{tabular}} & \multicolumn{1}{l|}{\%Reduction} & \multicolumn{1}{l|}{\begin{tabular}[c]{@{}l@{}}\%Accuracy on \\    random data\end{tabular}} \\ \hline
Full Dataset  & 97.35 & 0.00  & -  & {  87.41}  & 0.00  & -  \\ \hline
RHC & 90.03	& 95.06	& 93.14	& 64.34	& 90.93	& 82.52 \\ \hline
RHCKON  & {  96.02}  & 90.13  & 94.36   & 75.91   & 81.86    & 84.24   \\ \hline 
KONCW  & 96.70   & 74.05 & {  96.10}  & {  82.40}   & 69.07  & 84.10  \\ \hline CWKC & 95.72  & 79.00  & {  95.70}  & {  82.06}   & 78.14    & {  83.92} \\ \hline GHCIDR  & \textbf{96.83}  & \textbf{87.27}    & {  94.41}  & \textbf{83.96}  & \textbf{76.80} & 82.88  \\ \hline
\end{tabular}
\caption{Accuracy on reduced data, random data and \% reduction of MNIST and FMNIST datasets trained on FCN model for all four variants and RHC. }
\label{table:FCN}
\end{table*}

\begin{table*}[]
\centering
\begin{tabular}{|l|r|r|r|r|r|r|}
\hline
 & \multicolumn{6}{c|}{\textbf{VGG1}} \\ \hline
 & \multicolumn{3}{c|}{\textbf{MNIST}} & \multicolumn{3}{c|}{\textbf{FMNIST}} \\ \hline
Algorithm & \multicolumn{1}{l|}{\begin{tabular}[c]{@{}l@{}}\%Accuracy on\\    reduced data\end{tabular}} & \multicolumn{1}{l|}{\%Reduction} & \multicolumn{1}{l|}{\begin{tabular}[c]{@{}l@{}}\%Accuracy on\\    random data\end{tabular}} & \multicolumn{1}{l|}{\begin{tabular}[c]{@{}l@{}}\%Accuracy on\\    reduced data\end{tabular}} & \multicolumn{1}{l|}{\%Reduction} & \multicolumn{1}{l|}{\begin{tabular}[c]{@{}l@{}}\%Accuracy on\\    random data\end{tabular}} \\ \hline
Full Dataset & 99.51 & 0.00 & - & 93.25 & 0.00 & - \\ \hline
RHC & 97.35	& 95.06	& 98.00 & 75.39 & 90.93 & 87.86 \\ \hline
RHCKON & {  99.14} & 90.13 & {  98.70} & 88.60 & 81.86 & 89.51 \\ \hline
KONCW & 99.46 & 74.05 & 99.12 & 90.96 & 69.07 & 90.18 \\ \hline
CWKC & 99.21 & 79.00 & {  98.98} & {  89.74} & 78.14 & 90.13 \\ \hline
GHCIDR & \textbf{99.35} & \textbf{87.27} & 98.16 & \textbf{91.66} & \textbf{76.80} & 90.28 \\ \hline
\end{tabular}
\caption{Accuracy on reduced data, random data and \% reduction of MNIST and FMNIST datasets trained on VGG1 model for all four variants and RHC. }
\label{table:VGG1}
\end{table*}

\section{Evaluation}
% Please add the following required packages to your document preamble:
% \usepackage[table,xcdraw]{xcolor}
% If you use beamer only pass "xcolor=table" option, i.e. \documentclass[xcolor=table]{beamer}

\subsection{Datasets}
\textbf{MNIST Dataset:} It is collection of 60,000 handwritten images with 10 classes. Dimensions of each image are 28x28.\\
\textbf{FMNIST Dataset:} It is collection of 60,000 clothes images with 10 classes. Dimension of each image are 28x28.\\
\textbf{CIFAR10 Dataset:} It is a dataset of 50,000 coloured images with 10 classes. Dimension of each image are 32x32x3. There is a larger version of this dataset with 100 classes called CIFAR100.

\subsection{Metrics}
\textbf{Accuracy:} We are using accuracy as a metric to compare the experimental results.
% % % Distance Metrics: We used various distance metrics for evaluating the distance between 2 images like`$\ell 1$norm and $\ell 2$ norm. It was found that L2 norm performs better than L1 the norm in all experiments
\subsection{Experimental Settings}
\begin{itemize}[noitemsep,,topsep=0pt]
    \item We experimented on two deep learning algorithms are FCN (Fully Connected Network), and VGG1. We used Nvidia P100 GPU with 12 GB RAM for dataset reduction and training.
    \item Four variants and the original RHC algorithm were applied on the training datasets to obtain the reduced dataset. 
    \item For FCN and VGG1, we used a batch size of 32 and 64, and epochs equal to 5 and 100, respectively.
    \item We sampled a \emph{random} subset of images whose size was equal to the reduced dataset. For example in Table~ \ref{table:FCN}, RHCKON on MNIST reduces dataset size by 90.13\% or samples 9.87\% of the dataset. We select 9.87\% of the points randomly in the random sampling baseline. We report the average accuracy on randomly sampled datasets was calculated by varying the random seed.
    \item We evaluated the testing datasets and reported the accuracy for models trained on reduced data and random data.

\end{itemize}
% model = Sequential()
% 	model.add(Conv2D(32, (3, 3), activation='relu', kernel_initializer='he_uniform', padding='same', input_shape=(32, 32, 3)))
% 	model.add(Conv2D(32, (3, 3), activation='relu', kernel_initializer='he_uniform', padding='same'))
% 	model.add(MaxPooling2D((2, 2)))
% 	model.add(Dropout(0.2))
% 	model.add(Conv2D(64, (3, 3), activation='relu', kernel_initializer='he_uniform', padding='same'))
% 	model.add(Conv2D(64, (3, 3), activation='relu', kernel_initializer='he_uniform', padding='same'))
% 	model.add(MaxPooling2D((2, 2)))
% 	model.add(Dropout(0.2))
% 	model.add(Conv2D(128, (3, 3), activation='relu', kernel_initializer='he_uniform', padding='same'))
% 	model.add(Conv2D(128, (3, 3), activation='relu', kernel_initializer='he_uniform', padding='same'))
% 	model.add(MaxPooling2D((2, 2)))
% 	model.add(Dropout(0.2))
% 	model.add(Flatten())
% 	model.add(Dense(128, activation='relu', kernel_initializer='he_uniform'))
% 	model.add(Dropout(0.2))
% 	model.add(Dense(10, activation='softmax'))

% model = Sequential()
% model.add(Dense(512, input_shape=(784,)))
% model.add(Activation('relu'))
% model.add(Dropout(0.2))
% model.add(Dense(512))
% model.add(Activation('relu'))
% model.add(Dropout(0.2))
% model.add(Dense(10))
% model.add(Activation('softmax'))
% model.compile(loss='categorical_crossentropy', optimizer='adam', metrics=['accuracy'])
% print(type(X_train))
% model.fit(X_train, Y_train,batch_size=32, epochs=5,verbose=1)

\subsection{Results}

From Table \ref{table:FCN}, we can observe that:
\begin{itemize}[noitemsep,,topsep=0pt]     
    \item Accuracy of full dataset of MNIST when trained on FCN is 97.35\%. When reduced with GHCIDR, it gave an accuracy of 96.83\% and a reduction of 87.27\%. 
    \item Accuracy of full dataset of Fashion-MNIST, on FCN is 87.41\%. When reduced with GHCIDR, it gave an accuracy of 83.96\% and a reduction of 76.80\%.
\end{itemize}
From Table \ref{table:VGG1}, we can observe that:
\begin{itemize}[noitemsep,,topsep=0pt]
    \item Accuracy of full dataset of MNIST on VGG1 is 99.51\%. When reduced with GHCIDR, it gave an accuracy of 99.35\% and a reduction of 87.27\%.
    \item Accuracy of full dataset of Fashion-MNIST on VGG1 is 93.25\%. When reduced with GHCIDR, it gave an accuracy of 91.66\% and a reduction of 76.80\%.
\end{itemize}
From Table \ref{table:CIFAR_VGG}, we can observe that:
\begin{itemize}[noitemsep,,topsep=0pt]
    \item Accuracy of full dataset of CIFAR10 on VGG1 is 82.87\%. When reduced with GHCIDR, it gave an accuracy of 81.10\% and a reduction of 32.34\%.

    % \item A reduction of \textbf{87.27\%} on MNIST with an accuracy of \textbf{99.35\%} on VGG1 model.
    % \item A reduction of \textbf{76.8\%} on fashion MNIST with an accuracy of \textbf{91.66\%} on VGG1 model.
    % \item A reduction of \textbf{32.34\%} on CIFAR10 with an accuracy of \textbf{81.1\%} on VGG1 model.
\end{itemize}
\begin{table*}[]
\centering
\begin{tabular}{|l|r|r|r|}
\hline
\multicolumn{4}{|c|}{\textbf{VGG1}} \\ \hline
\multicolumn{4}{|c|}{\textbf{CIFAR}} \\ \hline
Algorithm & \multicolumn{1}{l|}{\%Accuracy on reduced data} & \multicolumn{1}{l|}{\%Reduction} & \multicolumn{1}{l|}{\%Accuracy on random data} \\ \hline
Full Dataset & 82.87 & 0.00 & - \\ \hline
RHC	& 67.59	& 75.67 & 70.8 \\ \hline
RHCKON & 76.04 & 51.34 & 75.24 \\ \hline
KONCW & 77.80 & 47.33 & 77.93 \\ \hline
CWKC & {  76.24} & 42.37 & 78.26 \\ \hline
GHCIDR & \textbf{81.10} & \textbf{32.34} & {  79.57} \\ \hline
\end{tabular}
\caption{Accuracy on reduced data, random data and \% reduction of CIFAR10 datasets trained on VGG1 model for all four variants and RHC.}
\label{table:CIFAR_VGG}
\end{table*}
\subsection{Analysis}

All four approaches reduced the dataset efficiently and gave an accuracy similar to that of the full dataset. They all performed better than the baseline RHC in terms of accuracy. GHCIDR outperformed all the variants for every dataset in all experiments. For the FCN model on MNIST and FMNIST datasets, KONCW gave a higher accuracy than RHCKON and CWKC but with a reduced \% reduction. Because cluster weights in KONCW, ensuring that the selection of images was proportional to the size of clusters. For the same amount of reduction, random data also gave less accuracy than KONCW. GHCIDR gave more reduction than KONCW and more accuracy because it samples images evenly from the cluster. KONCW inculcates the advantages of boundary points which help to increase the classification but decreases the reduction. On the other hand, GHCIDR takes advantage of KONCW and CWKC and considers the geometrical aspect of the data distribution. This leads to increased overall accuracy while keeping high \% reduction for the GHCIDR algorithm. By tuning the GHCIDR parameters - $\alpha$ \& $\beta$, we can achieve more accuracy but at the cost of a lesser reduction rate. This leads to a tradeoff between accuracy and reduction rate.

\section{Conclusion and Future Work}
We proposed novel variations to the RHC algorithm in this work. The best performing approach called GHCIDR was based on the intuition that we can get representative and diverse samples by sampling ``throughout'' the volume of the cluster. GHCIDR performed better than the proposed approaches, random baseline and the original RHC. 
% Trying to reduce it along with almost similar accuracy would be a great deal, and in the future, we might try to take on this task and improve the overall efficiency of our algorithm.

In the future, we plan to compare this work with non-RHC algorithms like Light-weight coresets~\cite{bachem2018scalable}, and other sampling algorithms. And we also plan to try different distance metrics such as SSIM~\cite{bae2015novel} and image augmentation techniques to modify the current algorithm. We also plan to reduce Imagenet~\cite{deng2009imagenet} and similar large datasets using the proposed approaches.
\bibliography{main}
\bibliographystyle{icml2021}

\end{document}